\documentclass[10pt,twocolumn,letterpaper]{article}

\usepackage{iccv}
\usepackage{times}
\usepackage{epsfig}
\usepackage{graphicx}
\usepackage{amsmath}
\usepackage{amssymb}
\usepackage[pagebackref=true,breaklinks=true,letterpaper=true,colorlinks,bookmarks=false]{hyperref}

\usepackage[noend]{algpseudocode}
\usepackage{algorithm}
\usepackage{amsmath}
\usepackage{amssymb}
\usepackage{amsfonts}
\usepackage{booktabs}
\usepackage{multirow}
\usepackage{graphicx}         %
\usepackage{color}
\usepackage{cite}             %
\usepackage{comment}          %
\usepackage{soul}             %
\soulregister\cite7
\soulregister\ref7
\soulregister\pageref7
\usepackage{amsthm}
\usepackage{etoolbox}         %
\usepackage{url}
\usepackage{nth}              %
\usepackage{bm}               %
\usepackage{caption} 
\usepackage{subfig}

\usepackage{cleveref}                                      %

\DeclareMathOperator*{\argmin}{argmin}

\makeatletter
\let\OldStatex\Statex
\renewcommand{\Statex}[1][3]{%
  \setlength\@tempdima{\algorithmicindent}%
  \OldStatex\hskip\dimexpr#1\@tempdima\relax
}
\makeatother

\iccvfinalcopy %

\def\iccvPaperID{5536} %

\usepackage{algorithm}
\usepackage{graphicx}
\usepackage{threeparttable}
\usepackage{color}
\usepackage{colortbl}
\usepackage{multirow}
\usepackage{amsmath}
\usepackage{comment}
\usepackage{bm}
\usepackage{etoolbox}                                      %
\usepackage{bbm}
\usepackage{url}
\usepackage{nth}
\usepackage{cite}
\usepackage{balance}
\usepackage{courier}                                       %
\usepackage{booktabs}                                      %
\usepackage{listings}                                      %
\usepackage[]{algpseudocode}
\usepackage{upgreek}
\usepackage[normalem]{ulem}
\useunder{\uline}{\ul}{}
\usepackage[accsupp]{axessibility}  %

\algdef{SE}[DOWHILE]{Do}{doWhile}{\algorithmicdo}[1]{\algorithmicwhile\ #1}%

\graphicspath{{./figs/}}

\begin{document}

\pagestyle{plain} %

\title{
Towards Memory-Efficient Neural Networks\\ via Multi-Level \textit{in situ} Generation
}

\author{
Jiaqi Gu, Hanqing Zhu, Chenghao Feng, Mingjie Liu, Zixuan Jiang, Ray T. Chen, David Z. Pan\\
The University of Texas at Austin\\
{\tt\small \{jqgu, hqzhu, fengchenghao1996, jay\_liu, zixuan\}@utexas.edu, \{chen, dpan\}@ece.utexas.edu}
}

\maketitle
\begin{abstract}
\label{abstract}
Deep neural networks (DNN) have shown superior performance in a variety of tasks.
As they rapidly evolve, their escalating computation and memory demands make it challenging to deploy them on resource-constrained edge devices.
Though extensive efficient accelerator designs, from traditional electronics to emerging photonics, have been successfully demonstrated, they are still bottlenecked by expensive memory accesses due to tremendous gaps between the bandwidth/power/latency of electrical memory and computing cores.
Previous solutions fail to fully-leverage the ultra-fast computational speed of emerging DNN accelerators to break through the critical memory bound.
In this work, we propose a general and unified framework to trade expensive memory transactions with ultra-fast on-chip computations, directly translating to performance improvement.
We are the first to jointly explore the intrinsic correlations and bit-level redundancy within DNN kernels and propose a multi-level \textit{in situ} generation mechanism with mixed-precision bases to achieve on-the-fly recovery of high-resolution parameters with minimum hardware overhead.
Extensive experiments demonstrate that our proposed joint method can boost the memory efficiency by 10-20$\times$ with comparable accuracy over four state-of-the-art designs, when benchmarked on ResNet-18/DenseNet-121/MobileNetV2/V3 with various tasks.

\end{abstract}

\section{Introduction}
\label{sec:Introduction}
Deep neural networks (DNNs) have demonstrated record-breaking performance in a variety of intelligent tasks.
Modern DNN models and datasets keep growing rapidly, which demonstrate critical conflicts with resource-constrained applications.
Stringent constraints in efficiency, latency, and power in practical applications raise a surging need to develop more efficient computing solutions.

Extensive efficient neural network (NN) accelerators have been designed to support such domain-specific computations.
In electrical domain, hardware-efficient digital platforms have been demonstrated, e.g., Eyeriss~\cite{NN_ISSCC2016_Chen,NN_JSSC2017_Chen}, EIE~\cite{NN_ISCA2016_Han},  TPU~\cite{NN_ISCA2017_Jouppi}.
Due to the high efficiency of analog computing, electrical analog accelerators gain much momentum recently, e.g., ReRAM-crossbar-based matrix multiplication engines~\cite{NN_ISCA2016_Shafiee,NN_HPCA2017_Song,NN_DATE2020_Wang}.
As a promising substitute for electrical designs to continue Moore's law, optical computing provides order-of-magnitude higher efficiency than electrical counterparts.
In optical computing domain, photonic accelerators are proposed to provide considerably more efficient solutions to AI acceleration~\cite{NP_NATURE2017_Shen,NP_DATE2019_Liu,NP_DATE2020_Zokaee,NP_HPCA2020_Shiflett, NP_ASPDAC2019_Zhao, NP_HotChips2020_Ramey, NP_CLEO2020_Feng, NP_ASPDAC2020_Gu, NP_TCAD2020_Gu,NP_SciRep2017_Tait, NP_APR2020_Miscuglio, NP_DATE2021_Gu, NP_DATE2021_Gu2, NP_Nature2020_Wetzstein, NP_NaturePhotonics2021_Shastri}.

However, memory performance turns out to be the critical bottleneck since it fails to match the computing capability of emerging cores.
Especially for emerging accelerators, e.g., ReRAM-based and photonics-based engines, the enormous latency, power, and bandwidth gap between memory and computing engines severely prohibits the full utilization of their advanced computing power.

Previous efforts towards memory-efficient accelerator designs focus on weight quantization~\cite{NN_Arxiv2016_Zhou,NN_ECCV2016_Rastegari,NN_ICLR2016_Han}, pruning with sparsity exploration~\cite{NN_ICML2020_Li,NN_ICLR2016_Han,NN_NIPS2015_Han,NN_NIPS2016_Wen,NN_ECCV2018_Zhang,NN_PIEEE2020_Deng}, structured weight matrices~\cite{NN_MICRO2017_Ding,NN_ICCAD2017_Liao,NP_ASPDAC2020_Gu,NN_DATE2020_Wang}, slim architectures~\cite{NN_ICLR2017_Iandola,NN_CVPR2017_Chollet,NN_ICLR2020_Han}, better hardware scheduling~\cite{NN_MLSYS2019_Lym,NN_HPCA2020_Zhang}, low-rank approximation~\cite{NN_NIPS2014_Denton, NN_ICLR2016_Tai,NN_ICML2020_Li, NN_ISCA2020_Zhao,NN_CVPRW2020_Yang,NN_IJCAI2020_Xu}, etc.
However, limited research has been done to thoroughly investigate the intrinsic redundancy in CNN kernels.
It is in high demand to provide a unique memory optimization strategy that fully exploits the potentials of advanced ultra-fast AI acceleration platforms.

Therefore, in this work, we propose a unified framework that generalizes prior low-rank solutions for memory-efficient NN designs via a multi-level \textit{in situ} weight generation technique with mixed-precision quantization.
We are the first to jointly explore multi-level redundancy in channel, kernel, and bitwidth based on a strong intuition on the intrinsic correlations within convolutions.
A photonic case study of \textit{in situ} weight generator is presented to show how our method can help unleash the full power of emerging neuromorphic computing systems.
The main contributions of this work are as follows,
\begin{itemize}
    \item We explore the multi-level intrinsic correlation in CNNs and propose a unified framework that generalizes prior low-rank-based convolution designs for higher memory efficiency.
    \item We fully-leverage the ultra-fast execution speed of emerging accelerators and propose a hardware-aware multi-level \textit{in situ} generation to trade expensive memory access for much cheaper computations.
    \item We integrate a precision-preserving mixed-precision strategy to leverage the bit-level redundancy in multi-level bases for a larger design space exploration.
    \item Experiments and a photonic case study show that our proposed multi-level \textit{in situ} generation and mixed-precision techniques can save $\sim$97\% weight load latency and significantly reduce memory cost by 10-20$\times$ with competitive accuracy compared to prior methods, even on compact networks and complex tasks.
\end{itemize}

\section{Preliminary}
\label{sec:Background}

In this section, we give a brief introduction to the background knowledge and our motivation.

\subsection{Memory bottleneck in NN accelerator designs}
\label{sec:NNDesign}

\begin{figure}
    \centering
    \subfloat[]{\includegraphics[width=0.21\textwidth]{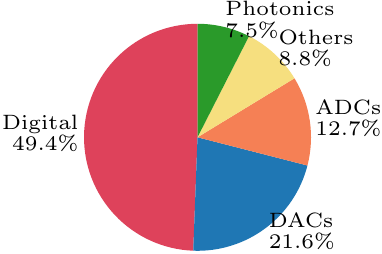}
    \label{fig:LightMatterPower}
    }
    \subfloat[]{\includegraphics[width=0.24\textwidth]{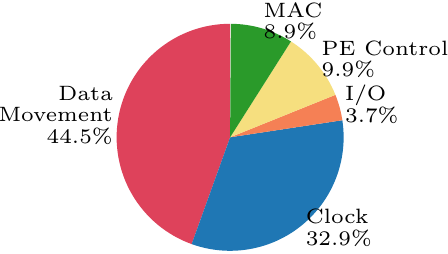}
    \label{fig:EyerissPower}
    }\\
    \vspace{-3pt}
    \subfloat[]{\includegraphics[width=0.165\textwidth]{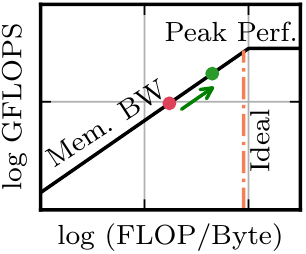}
    \label{fig:Roofline}
    }
    \hspace{20pt}
    \subfloat[]{\includegraphics[width=0.21\textwidth]{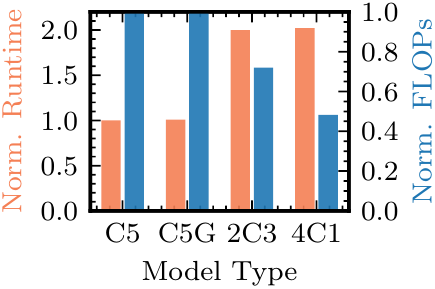}
    \label{fig:Runtime}
    }
    \caption{~\small
    Power breakdown of a silicon photonic accelerator \texttt{Mars}~\cite{NP_HotChips2020_Ramey, NP_JSTQE2020_Totovic} (a) and an electrical accelerator \texttt{Eyeriss}~\cite{NN_ISSCC2016_Chen} (b).
    The data movement (red) takes the most power for both.
    (c) Roofline model of emerging accelerators.
    Memory-bounded designs (red point) need to be improved to a better design (green point)
    (d) Normalized runtime and number of floating-point operations (FLOPs) among different convolution (Conv) types.
    C5 is 5$\times$5 Conv, C5G is 5$\times$5 Conv with low-rank decomposition, 2C3 is two cascade 3$\times$3 Conv, and 4C1 is four cascade 1$\times$3 Conv.
    }
    \label{fig:PowerBreakdown}
\end{figure}

Previous works have proposed extensive NN accelerator architectures to enable efficient DNN inference.
Recent emerging non-Von Neumann accelerators mainly focus on the innovation of the core matrix multiplication engine.
However, the computation speed and efficiency of the cores are no longer the bottlenecks of the overall system.
To prove this claim, Figure~\ref{fig:Runtime} shows that multiple cascaded small convolutional layers have less floating-point operations (FLOPs) than a single wide convolutional layer but have higher execution time due to lower parallelism and more memory transactions.
Hence, the expensive memory transaction and interconnect delay turn out to the pain point.

Most accelerators still rely on on-chip SRAMs and off-chip DRAMs to store/access weights, bringing serious challenges regarding the significant data movement cost.
First, the mismatch between memory and computing cores in terms of latency and bandwidth heavily limits the potential performance of modern accelerators, especially for ultra-fast optical accelerators.
Typical DRAM and SRAM has an access time of tens of nanoseconds, and the fastest SRAM runs at only 5 GHz.
However, for example, the computation is executed at the speed of light (picosecond-level delay) in optical NNs with massive parallelism and potentially over 100 GHz photo-detection rate~\cite{NP_NATURE2017_Shen,NP_Arxiv2020_Bernstein}.

Furthermore, data movement becomes the power bottleneck.
Figure~\ref{fig:LightMatterPower} shows the power breakdown on a recent photonic neural chip \texttt{Mars}~\cite{NP_HotChips2020_Ramey,NP_JSTQE2020_Totovic}.
The SRAM access dominates the total power consumption.
The same issue also exists in state-of-the-art (SOTA) electrical digital accelerators like famous \texttt{Eyeriss} \cite{NN_ISSCC2016_Chen,NN_JSSC2017_Chen} shown in Figure~\ref{fig:EyerissPower}.

Limited prior works have explicitly optimized memory cost for emerging accelerators by leveraging their ultra-fast computing speed.
Hence, a specialized memory-efficient NN design methodology to minimize data movement cost is exciting and essential to explore.

\subsection{Efficiency and accuracy trade-off}
\label{sec:MemoryComputeTradeoff}
Extensive works have been done to explore the NN design space for higher efficiency with less accuracy degradation.
Efficient neural architectures are designed with lightweight structures, e.g., depthwise separable convolution~\cite{NN_CVPR2017_Chollet}, blueprint convolution~\cite{NN_CVPR2020_Haase}, channel shuffling~\cite{NN_ICLR2017_Iandola}, etc.
Besides, network compression techniques are often utilized to explore the sparsity and redundancy of DNNs and trim the model size by pruning and quantization~\cite{NN_NIPS2015_Han,NN_ICLR2016_Han}.
Furthermore, low-rank decomposition~\cite{NN_ICML2020_Li, NN_ISCA2020_Zhao} is a widely adopted technique to reduce the number of parameters by approximating a weight matrix by two smaller matrices.
Also, structured neural networks\cite{NP_ASPDAC2020_Gu}\cite{NP_TCAD2020_Gu}\cite{NN_ICCAD2017_Liao} have been proposed to reduce memory cost with block-circulant matrix representation and Fourier-transform-based algorithm.

The above generic methods are applicable for emerging ultra-fast neuromorphic engines but do not fully leverage their powerful computing capability.
It will be interesting and promising to explore the intrinsic correlation in DNN weights and enable \textit{in situ} weight generation by the computing core itself to minimize data movement from memory.

\section{Proposed NN design methodology}
\label{sec:Method}
Motivated by prior work~\cite{NN_ICML2020_Li, NN_ISCA2020_Zhao, NN_CVPR2017_Chollet,NN_CVPR2020_Haase}, we focus on widely deployed convolutional neural networks (CNNs) to thoroughly explore their intrinsic multi-level redundancy for better efficiency.
We consider a 2-dimensional (2-D) convolutional kernel $\bm{W}\in\mathbb{R}^{C_o\times C_i \times k \times k}$ with $C_o$ kernels, $C_i$ input channels, and kernel sizes $k$.
Interestingly we observe intrinsic multi-level correlation within the kernel that we can leverage for memory compression.
This memory compression directly translates to latency/power improvement since convolutions have frequent weight access, whose memory cost is even higher than feature maps~\cite{NN_ISCA2016_Chen}.

\subsection{Multi-level weight generation}
\label{sec:WeightDecomposition}

\begin{figure}
    \centering
    \vspace{-5pt}
    \subfloat[]{\includegraphics[width=0.235\textwidth]{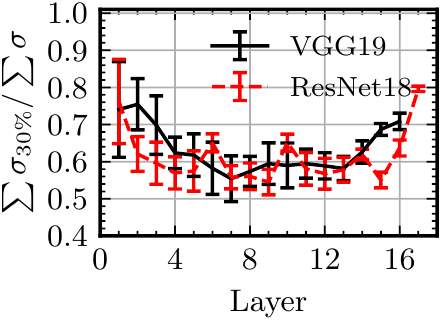}
    \label{fig:IntraKernelCorr}
    }
    \subfloat[]{\includegraphics[width=0.235\textwidth]{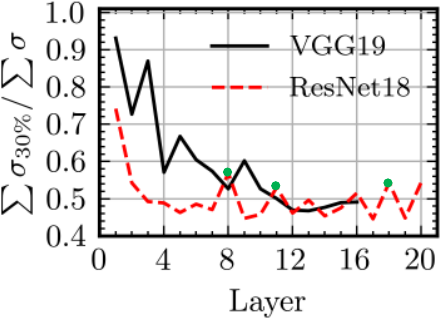}
    \label{fig:CrossKernelCorr}
    }
    \caption{~\small 
    Convolutional kernel correlations in ImageNet-pretrained models are shown by the proportion of the sum of the top 30\% singular values ($\sum\sigma_{30\%}$).
    (a) Intra-kernel correlations averaged on different kernels. 
    Error bars show the $\pm\sigma$ variance. 
    We skip 1$\times$1 Conv.
    (b) Cross-kernel correlations, where green dots are 1$\times$1 Conv.}
    \label{fig:KernelCorr}
    \vspace{-10pt}
\end{figure}

\subsubsection{Intra-kernel correlation}
\label{sec:IntraKernelCorr}
\begin{figure}
    \centering
    \includegraphics[width=0.49\textwidth]{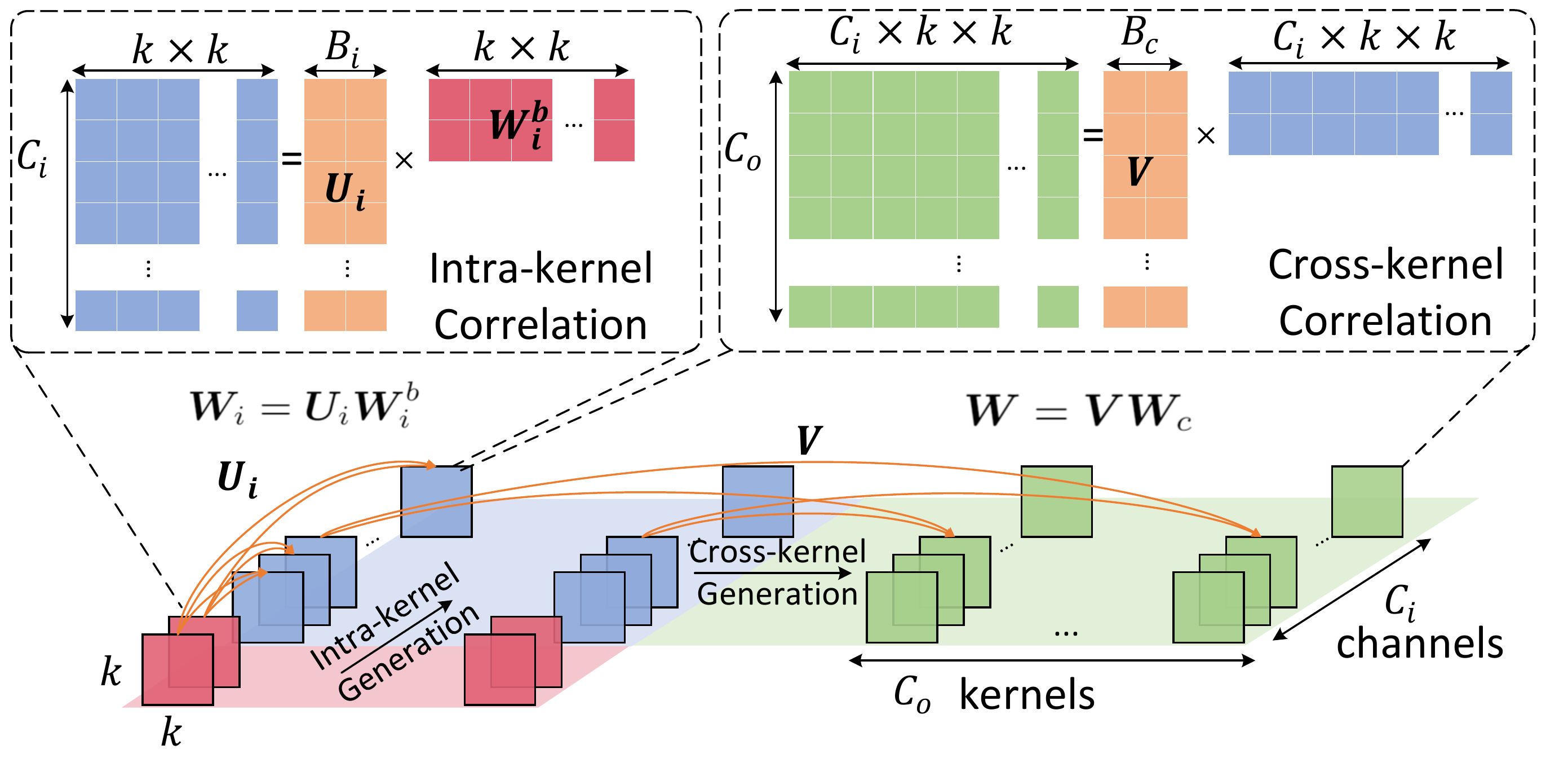}
    \vspace{-20pt}
    \caption{~\small
    Intra-kernel and cross-kernel generation.}
    \label{fig:LowrankDecomposition}
    \vspace{-5pt}
\end{figure}

We first explore the low-rank property among different channels of a kernel.
The $i$-th kernel $\bm{W}_i\in\mathbb{R}^{C_i\times k^2}$ can be treated as a matrix with $C_i$ row vectors with length $k^2$. 
From its singular values $\bm{\Sigma}=\texttt{SVD}(\bm{W}_i)=\texttt{diag}(\sigma_0,\sigma_1,\cdots)$, we observe relatively strong correlations between those column vectors since the first several major components $\sigma_{30\%}$ concentrates the majority of the total values.
Figure~\ref{fig:IntraKernelCorr} shows the intra-kernel low-rank property of modern CNNs.
Different layers tend to have different intra-kernel correlations, where shallower layers show higher correlations.
This provides us an opportunity to generate the $i$-th kernel $\bm{W}_i\in\mathbb{R}^{C_i\times k^2}$ using a low-dimensional \textit{channel basis} $\bm{W}^b_i\in\mathbb{R}^{B_i \times k^2}$ with a cardinality of $B_i<\min(C_i,k^2)$ and a corresponding coefficient matrix $\bm{U}_i\in\mathbb{R}^{C_i\times B_i}$. Figure~\ref{fig:LowrankDecomposition} visualizes the procedure for convolutions with a general matrix multiplication (GEMM) interpretation using the \textit{im2col} algorithm~\cite{NN_ICFHR2006_Chellapilla}.
This intra-kernel generation is formally expressed as.
\begin{equation}
    \small
    \label{eq:IntraKernelGeneration}
    \bm{W}_i=\bm{U}_i\bm{W}_i^b, \quad \forall i\in[C_o]
\end{equation}

Therefore, we reduce the parameter of the $i$-th kernel from $|\bm{W}_i|=C_ik^2$ to $|\bm{W}_i^b|+|\bm{U}_i|=B_ik^2+C_iB_i$.
Note that for 1$\times$1 convolution, we skip this intra-kernel generation and directly use all $C_i$ channels given the constraint $B_i<\min(C_i,1^2)$.

\subsubsection{Cross-kernel correlation}
\label{sec:CrossKernelCorr}
Furthermore, we explore the second-level correlation cross $C_o$ kernels.
We view the entire convolutional kernel $\bm{W}\in\mathbb{R}^{C_o\times(C_ik^2)}$ as a matrix with $C_o$ row vectors with length of $C_ik^2$.
Figure~\ref{fig:CrossKernelCorr} quantifies the correlation among different kernels.
Though it is slightly weaker than the intra-kernel correlation, it still brings another opportunity to further decompose the weight along another dimension.
Instead of generating $C_o$ kernels independently, we only generate a subset of kernels as our \textit{kernel basis} $\bm{W}_{c}=\{\bm{W}_i\in\mathbb{R}^{C_ik^2}, \forall i\in[B_c], B_c<\min(C_o,C_ik^2)\}$ using Eq.~\eqref{eq:IntraKernelGeneration}.
This generated kernel basis $\bm{W}_c$ is used to span the entire kernel together with another coefficient matrix $\bm{V}\in\mathbb{R}^{C_o\times B_c}$ as follows,
\begin{equation}
    \small
    \label{eq:CrossKernelGeneration}
    \bm{W}=\bm{V}\bm{W}_c=\bm{V}\{\bm{U}_{i}\bm{W}_{i}^b\}_{i\in[B_c]},
\end{equation}
If $B_c\geq\min(C_o,C_ik^2)$, we only consider intra-kernel correlation by setting $B_c=C_o$ without performing Equation~\eqref{eq:CrossKernelGeneration}.
After the proposed two-level generation, the parameter compression ratio is,
\begin{equation}
    \small
    \label{eq:CompressRatio}
    \begin{aligned}
    r=&\frac{|\bm{V}|+\sum_{i\in[B_c]}(|\bm{U}_i|+|\bm{W}_i^b|)}{|\bm{W}|}
    =\frac{\big(C_o+B_ik^2+C_iB_i\big)B_c}{C_oC_ik^2}.
    \end{aligned}
\end{equation}
The extra computation for \textit{in situ} kernel generation $\mathcal{O}(2B_cC_iB_ik^2+2C_oB_cC_ik^2)$ is marginal compared with the convolution itself $\mathcal{O}(2C_oC_ik^2HW)$, where $H$ and $W$ are output feature map sizes. 
Thus the runtime overhead is negligible, consistent with what we showed before in Figure~\ref{fig:Runtime}.
In this way, we successfully save expensive memory transactions with marginal computation overhead, which fully leverages the emerging accelerators' ultra-fast computing capability to mitigate the critical memory bound.

\subsection{Augmented mixed-precision generation}
\label{sec:BitLevelGeneration}
Besides the weight correlation that explores parameter-level reduction, we further explore the bit-level redundancy with mixed-precision bases.
Modern NN accelerator designs, especially emerging analog engines, prefer to use low-bit weights to reduce memory access latency and simplify the control circuitry complexity~\cite{NN_Arxiv2016_Zhou,NN_ECCV2016_Rastegari,NP_NATURE2017_Shen,NP_DATE2020_Gu,NN_DAC2020_Zheng}.
In this section, we utilize the precision preserving feature of analog engines and propose an augmented mixed-precision generation strategy to recover high-precision weights with low-bitwidth basis and coefficients.

We assume the bitwidths for $\bm{W}^b_i$, $\bm{U}_i$, and $\bm{V}$ are $q_b$, $q_u$, and $q_v$, respectively.
The first-level intra-kernel generation is capable of generating $\bm{W}_c\in\mathbb{R}^{B_c\times(C_ik^2)}$ with at most $(2^{q_b}-1)(2^{q_u}-1)B_i+1$ possible distinct values, which corresponds to a bitwidth upper bound $\sup(q_c)=(q_b+q_u+\log_2{B_i})$.
Unlike digital cores, this precision can be maintained by the direct cascade of two analog tensor units without resolution loss caused by the analog-to-digital conversion.
Then, the cross-kernel generator will output $\bm{W}$ with an equivalent bitwidth $\sup(q)=(q_v+\sup(q_c)+\log_2{B_o})$ that can also be preserved in the matrix multiplication unit.
The advantages are clear that our method enables the weight generator to be completely in the analog domain to recover a high-precision, i.e., $q>q_b,q_u,q_v$, weight matrix using low-precision basis and coefficient matrices.
The memory compression ratio $r_m$ is thus calculated as,
\begin{equation}
    \small
    \label{eq:MemCompressionRatio}
    \begin{aligned}
    r_m&=\frac{\sum_{i\in[B_c]}\big(q_b|\bm{W_i}^b|+q_u|\bm{U}_i|\big)+q_v|\bm{V}|}{q_w|\bm{W}|}\\
    &=\frac{B_cB_ik^2q_b+B_cC_iB_iq_u+C_oB_cq_v}{C_oC_ik^2q_w}.
    \end{aligned}
\end{equation}
Hence, given a target $q_w$, we can explore fine-grained mixed-precision settings of $q_b$, $q_u$, and $q_v$ to further cut down the memory cost in the bit-level, which is an orthogonal technique to the above parameter-level counterparts.

\subsection{Training with \textit{in situ} weight generation}
\label{sec:TrainingGenerator}
\begin{algorithm}[tb]
\caption{Training with \textit{in situ} generation}
\label{alg:Training}
\begin{algorithmic}[1]
\small
\Require A pretrained teacher $\widehat{\mathcal{M}}$ with weights $\widehat{\bm{W}}$, a student model $\mathcal{M}$ with $\bm{W}^b_i$, $\bm{U}_i$, and $\bm{V}$,
mixed-precision bitwidths $q_b$, $q_u$, and $q_v$,
training dataset $\mathcal{D}^{trn}$, total iterations $T$, initial step size $\eta^0$;
\Ensure Converged student model;
\State $\text{Step 1:~} \ell_2 \text{~Initialization from the teacher model}$
\State $\bm{W}^b_i, \bm{U}_i, \bm{V}\gets\argmin~\|\widehat{\bm{W}}-\bm{V}\{\bm{U}_i\bm{W}_i^b\}_{i\in[B_c]}\|_2^2$
\State $\text{Step 2: Quantization-aware knowledge distillation}$
\For{$t\gets 0\cdots T-1$}
\State Randomly sample a mini-batch $\mathcal{I}^t$ from $\mathcal{D}^{trn}$
\State $\bm{U}_i^{t+1} \gets \bm{U}_i^{t} - \eta^t \nabla_{\bm{U}_i}(\mathcal{L}_{KD}+\lambda\mathcal{L}_{ort}),~\forall i\in[B_c]$
\State $\bm{W}_i^{b,t+1} \gets \bm{W}_i^{b,t} - \eta^t \nabla_{\bm{W}^b_i}(\mathcal{L}_{KD}+\lambda\mathcal{L}_{ort}),~ \forall i\in[B_c]$
\State $\bm{V}^{t+1} \gets \bm{V}^{t} - \eta^t \nabla_{\bm{V}_i}(\mathcal{L}_{KD}+\lambda\mathcal{L}_{ort})$
\State $\eta^{t+1}=\texttt{Update}(\eta^t)$ \Comment{Step size decay}
\EndFor
\end{algorithmic}
\end{algorithm}
Our main target is to reduce memory cost with acceptable accuracy loss.
Now we introduce how to optimize the designed CNN with \textit{in situ} generators such that the desired accuracy can be achieved.
We adopt a two-stage quantization-aware knowledge distillation to train our proposed NN, described in Alg.~\ref{alg:Training}.
Firstly, we obtain a pre-trained full-precision model without \textit{in situ} generation as our teacher model $\widehat{\mathcal{M}}$ whose weight matrix is denoted as $\widehat{\bm{W}}$.
Our low-rank mixed-precision model is the corresponding student model $\mathcal{M}$ whose weight matrix $\bm{W}$ is generated by quantized $\bm{W}_i^b$, $\bm{U}_i$, and $\bm{V}$.
A differentiable quantizer~\cite{NN_Arxiv2016_Zhou} is used in our quantization-aware training.
For simplicity, we omit the quantization notation for quantized $\bm{W}_i^b$, $\bm{U}_i$, and $\bm{V}$ if mixed-precision quantization is used.
Then we let the student mimic the teacher using a two-stage training algorithm.
First, we solve the following problem to project the teacher model onto the student parameter space by minimizing their $\ell_2$ distance,
\begin{equation}
    \small
    \label{eq:Projection}
    \min~\|\widehat{\mathcal{M}}(\widehat{\bm{W}})-\mathcal{M}(\bm{W})\|_2^2\approx\|\widehat{\bm{W}}-\bm{V}\{\bm{U}_i\bm{W}_i^b\}_{i\in[B_c]}\|_2^2.
\end{equation}
Given the smoothness of $\mathcal{M}$ and $\widehat{\mathcal{M}}$, the above $\ell_2$ distance can be approximated by the first-order term of its Taylor expansion.
This $\ell_2$ distance-based subspace projection is an effective and efficient initialization method for the student model.
Then we try to find local optima in the low-rank space starting from this projected solution point.
Therefore, in the second stage, we train the student model with knowledge distillation~\cite{NN_Arxiv2015_Hinton} as,
\begin{equation}
    \small
    \label{eq:Formulation}
    \begin{aligned}
    \min ~&\mathcal{L}_{KD}=\beta T^2\mathcal{D}_{KL}(q_T,p_T)+(1-\beta)H(q,p_{T=1}), \\
    \text{s.t.}~~& p_T = \frac{\exp(\frac{\mathcal{M}(\bm{W})}{T})}{\sum\exp(\frac{\mathcal{M}(\bm{W})}{T})}, q_T = \frac{\exp(\frac{\widehat{\mathcal{M}}(\widehat{\bm{W}})}{T})}{\sum\exp(\frac{\widehat{\mathcal{M}}(\widehat{\bm{W}})}{T})},  \\
    & \bm{W}=\bm{V}\{\bm{U}_i\bm{W}_i^b\}_{i\in[B_c]}, \\
    & 0 < B_i <\min(C_i, k^2), B_i\in \mathbb{Z}\\
    & 0< B_c < \min(C_o, C_ik^2), B_i\in \mathbb{Z},
    \end{aligned}
\end{equation}
where $\mathcal{M}(\bm{W})$ is the output logits, $\mathcal{D}_{KL}$ is the Kullback–Leibler divergence between two probability distributions, $H(\cdot,\cdot)$ is the cross entropy, $q$ is the ground truth distribution, $T$ and $\beta$ are hyper-parameters controlling the smoothness.
This training method~\cite{NN_Arxiv2015_Hinton} can distill the representability of the high-rank full-precision model to our low-rank quantized student.

However, we notice that once the basis and coefficient matrices have a deficient row-rank or column-rank, the spanning subspace of the generated matrix will become too small to approximate the original full-rank matrix. 
Therefore, to maximize the rank of the spanned weight matrix, we set a row orthonormality constraint to the basis $\bm{W}^b_i$ and a column orthogonality constraint to the coefficient matrices.
This constraint can be relaxed using penalty methods as a multi-level orthogonal regularization term $\mathcal{L}_{ort}$ as follows,
\begin{equation}
    \small
    \label{eq:Orthogonality}
    \begin{aligned}
&\sum_{i=1}^{B_c}\!\Big(\|\bm{W}^b_i(\bm{W}^b_i)^T\!-\!\bm{I}\|_2^2\!+\!\|\tilde{\bm{U}}_i^T\tilde{\bm{U}}\!-\!\bm{I}\|_2^2\Big)\!+\!\|\tilde{\bm{V}}^T\tilde{\bm{V}}\!-\!\bm{I}\|_2^2,\\
    &\tilde{\bm{U}}_i=\begin{pmatrix}
    \frac{u_0}{\|u_0\|_2^2} & \!\!\!\!\cdots\!\!\!\! & \frac{u_0}{\|u_{B_i-1}\|_2^2}
    \end{pmatrix},
    \tilde{\bm{V}}=\begin{pmatrix}
    \frac{v_0}{\|v_0\|_2^2} & \!\!\!\!\cdots\!\!\!\! & \frac{v_0}{\|v_{B_c-1}\|_2^2}
    \end{pmatrix}.
    \end{aligned}
\end{equation}
Equation~\eqref{eq:Orthogonality} is a generalization to a previous single-level penalty~\cite{NN_CVPR2020_Haase, NN_CVPRW2020_Yang} and exerts a soft constraint to multi-level correlations such that the spanning space will not collapse to a low-dimensional subspace.
Therefore, the overall loss function is $\mathcal{L}=\mathcal{L}_{KD}+\lambda \mathcal{L}_{ort}$.

\subsection{Case study: silicon photonics implementation}
\label{sec:HardwareRealization}
\begin{figure}
    \centering
    \includegraphics[width=0.49\textwidth]{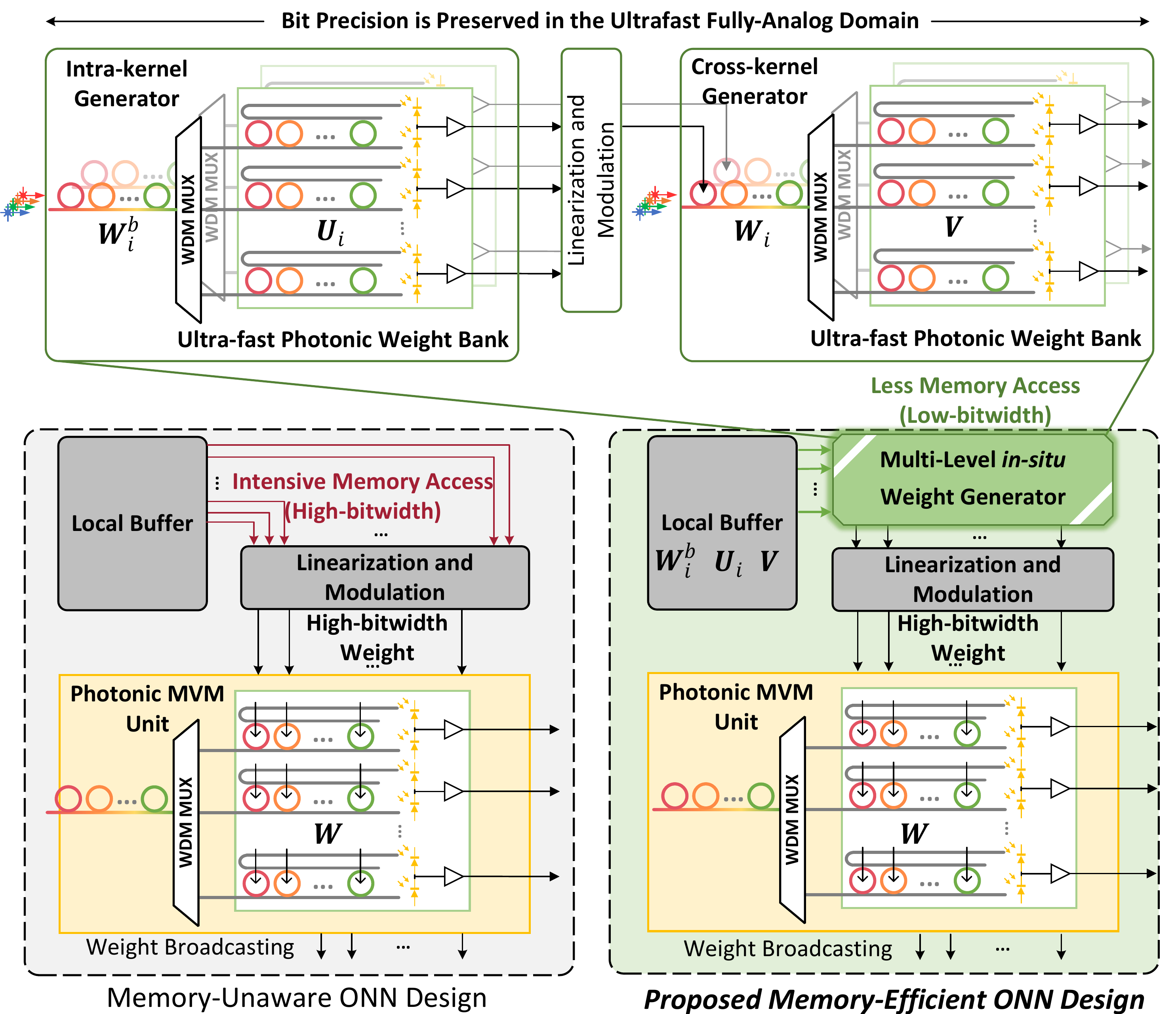}
    \caption{~\small
    Photonic implementation of \textit{in situ} weight generator and peripheral structures.
    \textit{Left bottom}}
    \label{fig:PhotonicImplementation}
\end{figure}
\begin{figure*}
    \centering
    \vspace{-10pt}
    \subfloat[]{\includegraphics[width=0.22\textwidth]{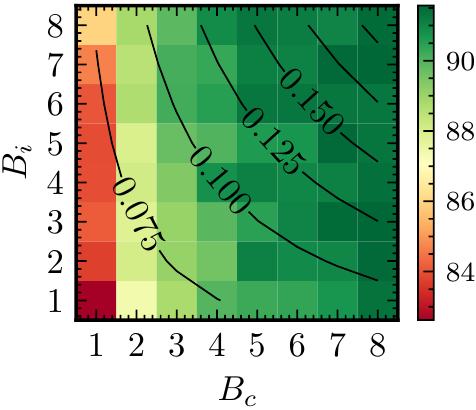}
    \label{fig:LowrankSearchFMNIST}
    }
    \subfloat[]{\includegraphics[width=0.753\textwidth]{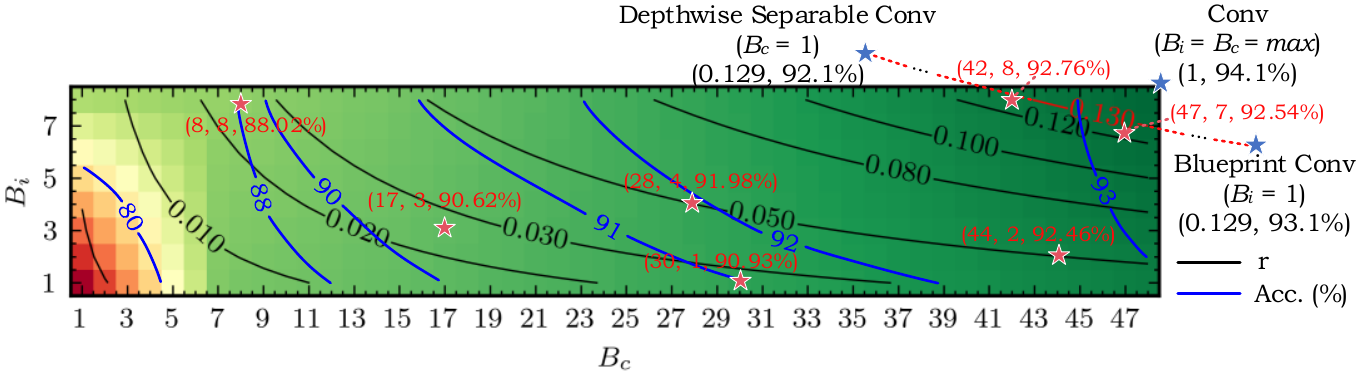}
    \label{fig:LowrankSearchCIFAR}
    }
    \caption{~\small
    (a) Accuracy (color) and compression ratio (contour) of the customized 3-layer CNN on FashionMNIST~\cite{NN_FashionMNIST2017} with various $B_i$ and $B_c$ (92.14\% Acc. for the original Conv).
    (b) Accuracy (blue contour) and compression ratio $r$ (black contour) for ResNet-18 on CIFAR-10.
    Red stars are representative settings of our method.
    Blue stars show previous designs.
    }
    \label{fig:LowrankSearch}
    \vspace{-10pt}
\end{figure*}
We showcase a photonic implementation of the proposed \textit{in situ} weight generator in Figure~\ref{fig:PhotonicImplementation}.
We focus on a SOTA design based on micro-ring resonators~\cite{NP_SciRep2017_Tait}.
Other accelerators can also benefit from our method as long as the multi-level correlation and precision preserving properties hold.

After loading the lightweight basis and coefficient matrices from the local electrical buffer, two cascaded ultra-fast optical weight banks will achieve the first-level and second-level generation to obtain the final weights $\bm{W}$.
Without intermediate storage, the analog weights are directly broadcast to all photonic tensor units via ultra-low-power optical interconnects~\cite{NP_Arxiv2020_Bernstein} to perform the primary operation, e.g., convolution.
Compared with the memory-agnostic design, which requires massive and frequent weight loading, our proposed design can effectively cut down memory footprint and access latency.
Consider a 16-bit ($q_w$=16) kernel $\bm{W}\in\mathbb{R}^{128\times128\times3\times3}$ and a setting ($B_i,B_c,q_b,q_u,q_v$)=(2,40,4,4,4) implemented by micro-rings of diameter $R$=20~$\upmu$m, the extra latency introduced by \textit{in situ} generator is as follows,
\begin{equation}
    \small
    \label{eq:LatencyOverhead}
    \begin{aligned}
    \tau_{gen}&=(\tau_{DAC}+\tau_{mod}+\tau_{prop1}+\tau_{oe})+(\tau_{mod}+\tau_{prop2}+\tau_{oe})\\
    &\approx\tau_{DAC}+2\times(\tau_{mod}+\tau_{oe})+\frac{4B_iR}{c}+\frac{4B_cR}{c}\\
    &\approx 400~\text{ps} + 2\times(50~\text{ps} + 10~\text{ps})+25.2~\text{ps}=545.2~\text{ps}\\
    &\ll \frac{2(1-r_m)|\bm{W}|}{BW_{SRAM}}\!\approx\!\frac{(1-0.0272)\times288~\text{KB}}{34~\text{GB/s}}= 7.9~\text{$\upmu$s},
    \end{aligned}
\end{equation}
where $\tau_{DAC}$ is the latency for 10 Gb/s digital-to-analog converter, $\tau_{mod}$ is the device modulation delay, $\tau_{prop}$ is the photonic weight bank propagation delay, $\tau_{oe}$ is the optical-to-electrical conversion delay for layer cascade, $c$ is the light speed, and $BW_{SRAM}$ is the SRAM bandwidth~\cite{NN_ISCA2017_Jouppi}.
The generator saves 7.9 $\upmu$s latency ($>$97\% of total weight load latency) with merely 545.2 $\text{ps}$ weight generation latency overhead.
Given $\sim$50\% of total latency is consumed by kernel loading~\cite{NN_ISCA2016_Chen}, our weight generation leads to at least 2$\times$ overall speedup.
More speedup can be expected if activation quantization is further applied.
In terms of power, our method can achieve significant energy reduction since we save $(1-r_m)\approx97\%$ weight loading and replace all high-resolution DACs with $(1-r)\approx89\%$ fewer low-bit DACs~\cite{HW_TCAS2011_Saberi} (power is exponential to bitwidth), which account for most power as shown in Figure~\ref{fig:LightMatterPower}. 

We further perform quantitative evaluation on a neuromorphic simulator MNSIM-2.0.
On ResNet-18/ImageNet, compared with 8-bit \texttt{BSConv}, our method reduces the overall latency from 56.46 ms to 41.11 ms (\textbf{27.2\%}$\downarrow$), reduces the overall energy from 25.77 mJ to 3.69 mJ (\textbf{85.7\%}$\downarrow$), and improves energy-delay-product by \textbf{9.6$\times$}.

\vspace{-.05in}
\section{Experiments}
\label{sec:ExperimentalResults}
In this section, we first conduct ablation experiments on the proposed techniques and compare our method with prior efficient designs in memory cost and accuracy.

\subsection{Dataset}
\label{sec:Dataset}
Our ablation and comparison experiments are based on FashionMNIST~\cite{NN_FashionMNIST2017}, CIFAR-10~\cite{NN_cifar2009}, and CIFAR-100.
We also test on more tasks including SVHN~\cite{NN_NIPS2011_Netzer}, TinyImagetNet-200~\cite{NN_CVPR2009_Deng}, StanfordDogs-120~\cite{NN_CVPR2011_Khosla} and StandfordCars-196~\cite{NN_3DRR2013_Krause} for fine-grained classification.

\subsection{Neural network architectures}
\label{sec:NNArchitectures}
We first use a customized 3-layer CNN as a toy example to do multi-level correlation exploration on FashionMNIST, whose settings are (C32K5S2-C32K5S1-C32K5S1-AvgPool3-FC10), where C32K5S2 is a 5$\times$5 convolution with 32 kernels and stride 2, AvgPool3 is an average pooling layer with output size 3$\times$3, and FC10 means the output linear layer.
BatchNorm and ReLU activation are used between convolutional layers.
Then, the rest ablation experiments and comparison experiments are based on ResNet-18~\footnote{\url{https://github.com/kuangliu/pytorch-cifar}}~\cite{NN_CVPR2016_He}, DenseNet-121~\footnote{\url{https://github.com/gpleiss/efficient_densenet_pytorch}}~\cite{NN_CVPR2017_Huang}, and MobileNetV2~\cite{NN_CVPR2018_Sandler}, which are adapted to CIFAR-10/100.

\subsection{Training settings}
\label{sec:TrainingSettings}
We train all models for 200 epochs using RAdam~\cite{NN_ICLR2020_Liu} optimizer with an initial learning rate of 0.002, an exponential decay rate of 0.98 per epoch, and a weight decay of 5e-4.
On CIFAR-10/100, images are augmented by random horizontal flips and random crops with 4 paddings.
On TinyImageNet, StanfordDogs-120, and StanfordCars-196, additional color jitter is added.
Mini-batch sizes are 64, 128, 64, and 64 for our 3-layer CNN, ResNet-18, DenseNet-121, and MobileNetV2, respectively.

\subsection{Ablation: multi-level correlation exploration}
\label{sec:LowrankExp}
To explore the impact of the multi-level basis cardinality $B_i$ and $B_c$ on the parameter count and accuracy, we first perform a grid search on FashionMNIST with our customized 3-layer CNN, shown in Figure~\ref{fig:LowrankSearchFMNIST}.
In terms of parameter compression ratio $r$, $B_c$ shows a stronger impact than $B_i$ since $r\propto B_c$ while $B_i$ only partially contributes to $r$.
For test accuracy, generally larger $B_i$ and $B_c$ lead to higher accuracy.
However, the accuracy is much more sensitive to $B_c$ than $B_i$, where we find a great opportunity to minimize memory cost with a small accuracy drop.
Therefore, we conclude a heuristic design guidance that a small $B_i$ and medium $B_c$ leads to sweet points.
We further validate it on CIFAR-10 with ResNet-18, whose contours are shown in
Figure~\ref{fig:LowrankSearchCIFAR}.
In the design space exploration, we also plot full-rank Conv, depthwise separable Conv~\cite{NN_CVPR2016_He}, and blueprint Conv~\cite{NN_CVPR2020_Haase} as our special cases.
The blueprint Conv can be generalized by our method once $B_i$=1 and $B_c$=max.
To some extent, separable Conv can also be generalized by setting $B_i$=max and $B_c$=1 while using different $\bm{V}$ for different input channels.
Note that sharing $\bm{V}$ across channels is the key-point for our efficiency superiority.
With the concluded design guidance, we indeed can quickly find design points that outperform the above prior works in memory efficiency with comparable accuracy, e.g., ($B_i$=2, $B_c$=44).
Note that we assume a global ($B_i$, $B_c$) setting for all layers, while layer-specific cardinalities can be an interesting future topic to push towards the Pareto front.

\subsection{Ablation: multi-level orthogonality regularization}
\label{sec:OrthogonalExp}
\begin{figure}
    \centering
    \includegraphics[width=0.4\textwidth]{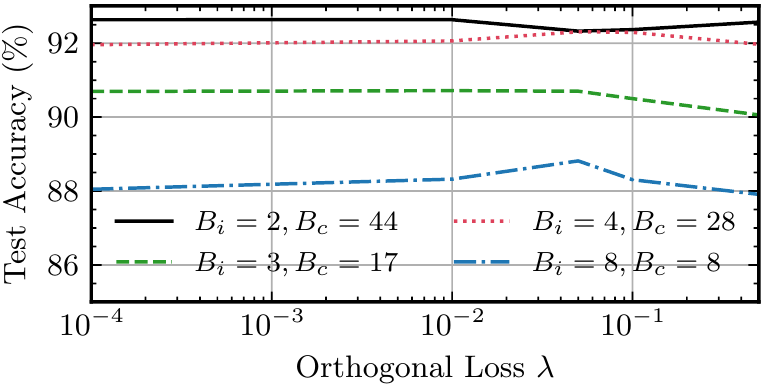}
    \caption{~\small
    Exploration on different orthogonal regularization weights with ResNet-18 on CIFAR-10\cite{NN_cifar2009}.
    }
    \label{fig:OrthogonalExp}
    \vspace{-6pt}
\end{figure}

\vspace{-3pt}
Several representative ($B_c$,$B_i$) pairs are evaluated on ResNet-18 CIFAR-10 with various regularization weights $\lambda$.
Figure~\ref{fig:OrthogonalExp} reveals that the model performance can be consistently improved by 0.5\%-1\% with proper $\lambda$ values ($0.01\sim0.05$).
This shows that the proposed multi-level orthogonal penalty term can encourage the spanned kernel to be as high-rank as possible with augmented representability.

\begin{table}[]
\centering
\resizebox{.48\textwidth}{!}{
\begin{tabular}{|l|c|c|c|c|c|c|c|c|}
\hline
\multirow{3}{*}{} & \multicolumn{4}{c|}{Param Ratio $r$=0.025}                                                  & \multicolumn{4}{c|}{Param Ratio $r$=0.05}                                                   \\ \cline{2-9}
                  & $B_i$                 & $B_c$               & $B_i$                 & $B_c$               & $B_i$                 & $B_c$               & $B_i$                 & $B_c$               \\ \cline{2-9}
                  & 3                  & 17               & 8                  & 8                & 2                  & 44               & 4                  & 28               \\ \hline\hline
Baseline              & \multicolumn{2}{c|}{90.62\%}          & \multicolumn{2}{c|}{88.02\%}          & \multicolumn{2}{c|}{92.46\%}          & \multicolumn{2}{c|}{91.98\%}          \\
Ortho Reg            & \multicolumn{2}{c|}{90.82\%}          & \multicolumn{2}{c|}{88.52\%}          & \multicolumn{2}{c|}{92.88\%}          & \multicolumn{2}{c|}{92.32\%}          \\
SVD Init               & \multicolumn{2}{c|}{91.32\%}          & \multicolumn{2}{c|}{88.10\%}          & \multicolumn{2}{c|}{93.05\%}          & \multicolumn{2}{c|}{92.80\%}          \\
$\ell_2$ Init                & \multicolumn{2}{c|}{91.32\%}          & \multicolumn{2}{c|}{88.85\%}          & \multicolumn{2}{c|}{93.18\%} & \multicolumn{2}{c|}{92.75\%}          \\
$\ell_2$+Ortho          & \multicolumn{2}{c|}{91.40\%}          & \multicolumn{2}{c|}{88.65\%}          & \multicolumn{2}{c|}{93.17\%}          & \multicolumn{2}{c|}{92.93\%}          \\
$\ell_2$+Ortho+KD       & \multicolumn{2}{c|}{\textbf{91.52\%}} & \multicolumn{2}{c|}{\textbf{88.96\%}} & \multicolumn{2}{c|}{\textbf{93.29\%}}          & \multicolumn{2}{c|}{\textbf{93.19\%}} \\ \hline
\end{tabular}
}
\caption{~\small
Accuracy evaluation on orthogonal regularization (\textit{Ortho}), initialization ($\ell_2$ and \textit{SVD}), and knowledge distillation (\textit{KD}).
ResNet-18 is evaluated on CIFAR-10.}
\label{tab:AblationCompare}
\end{table}

\subsection{Ablation: initialization and distillation}
\label{sec:Initialization}
We further evaluate different combinations of the proposed $\ell_2$ initialization and knowledge distillation with representative ($B_i,B_c$) pairs in Table.~\ref{tab:AblationCompare}.
In our $\ell_2$ initialization, we optimize Equation~\eqref{eq:Projection} using RAdam~\cite{NN_ICLR2020_Liu} for 3k iterations with lr=2e-2.
We first compare with a traditional truncated singular value decomposition (SVD) based method~\cite{NN_NIPS2014_Denton, NN_CVPRW2020_Yang}.
Both methods benefit accuracy while our $\ell_2$ initialization demonstrates better results.
With orthogonality penalty and knowledge distillation ($\beta$=0.9, $T$=3), our method achieves the highest accuracy.
In conclusion, a good initialization and knowledge from the teacher are critical to the accuracy of the student model.

\subsection{Ablation: mixed-precision bases exploration}
\label{sec:MixedPrecisionExp}
\begin{figure}
\vspace{-6pt}
    \centering
    \includegraphics[width=0.38\textwidth]{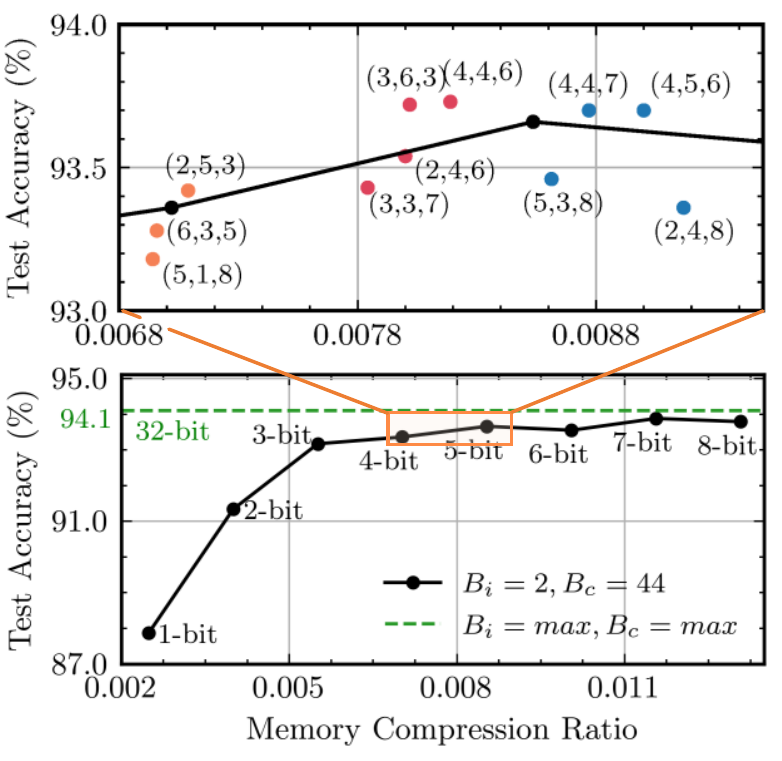}
    \caption{~\small
    Accuracy and memory compression ratio contour of ResNet-18 on CIFAR-10 with mixed-precision quantization ($q_b,q_u,q_v$).
    Black dots show $q_b$=$q_u$=$q_v$.
    }
    \label{fig:QuantCompare}
\end{figure}
\begin{table*}[]
\centering
\resizebox{0.9\textwidth}{!}{%
\begin{tabular}{|l|c|c|c|c|c|c|}
\hline
\multirow{2}{*}{}                 & \multicolumn{3}{c|}{CIFAR-10}     & \multicolumn{3}{c|}{CIFAR-100}    \\ \cline{2-7}
                                  & Param Ratio & Mem Ratio & Acc     & Param Ratio & Mem Ratio & Acc     \\ \hline\hline
ResNet-18 (\texttt{Conv})~\cite{NN_CVPR2016_He}                  & 1.0000      & 1.0000    & 94.10\% & 1.0000      & 1.0000    & 73.53\% \\\hline
ResNet-18 (\texttt{DSConv})~\cite{NN_CVPR2017_Chollet}                & 0.1287      & 0.1287    & 92.10\% & 0.1323      & 0.1323    & 68.65\% \\
ResNet-18 (\texttt{PENNI} d=2)~\cite{NN_ICML2020_Li}             & 0.2352      & 0.2352    & 92.77\% & 0.2383      & 0.2383    & 70.14\% \\
ResNet-18 (\texttt{BSConv})~\cite{NN_CVPR2020_Haase}                & 0.1291      & 0.1291    & 93.10\% & 0.1327      & 0.1327    & 71.11\% \\
ResNet-18 (\texttt{CirCNN} k=4)~\cite{NN_MICRO2017_Ding}                & 0.2510      & 0.2510    & 92.16\% & 0.2541      & 0.2541    & 67.93\% \\
ResNet-18 (\texttt{Ours}-2-44-32-32-32)    & 0.0497      & 0.0497    & 93.29\% & 0.0536      & 0.0536    & 70.85\% \\
ResNet-18 (\texttt{Ours}-2-44-8-8-8)       & 0.0497      & 0.0131    & \textbf{93.79\%} & 0.0536      & 0.0140    & 71.05\% \\
ResNet-18 (\texttt{Ours}-2-44-3-6-3)       & 0.0497      & \textbf{0.0080}    & 93.72\% & 0.0536      & \textbf{0.0090}    & \textbf{71.47\%} \\ \hline\hline
DenseNet-121 (\texttt{Conv})~\cite{NN_CVPR2017_Huang}               & 1.0000      & 1.0000    & 94.69\% & 1.0000      & 1.0000    & 76.51\% \\\hline
DenseNet-121 (\texttt{DSConv})~\cite{NN_CVPR2017_Chollet}             & 0.7362      & 0.7362    & 93.81\% & 0.7396      & 0.7396    & 74.35\% \\
DenseNet-121 (\texttt{PENNI} d=2)~\cite{NN_ICML2020_Li}          & 0.7608      & 0.7608    & 94.32\% & 0.7640      & 0.7640    & 75.26\% \\
DenseNet-121 (\texttt{BSConv})~\cite{NN_CVPR2020_Haase}             & 0.7291      & 0.7291    & 94.24\% & 0.7326      & 0.7326    & \textbf{75.79\%} \\
DenseNet-121 (\texttt{CirCNN} k=4)~\cite{NN_MICRO2017_Ding}             & 0.2601      & 0.2601    & 92.86\% & 0.2698      & 0.2698    & 72.45\% \\
DenseNet-121 (\texttt{Ours}-1-25-32-32-32) & 0.1986      & 0.1986    & \textbf{94.89\%} & 0.2091      & 0.2091    & 75.09\% \\
DenseNet-121 (\texttt{Ours}-1-25-8-8-8)    & 0.1986      & 0.0587    & 94.78\% & 0.2091      & 0.0612    & 75.59\% \\
DenseNet-121 (\texttt{Ours}-1-25-4-6-6)    & 0.1986      & \textbf{0.0395}    & 94.68\% & 0.2091      & \textbf{0.0422}    & 75.05\% \\ \hline
\end{tabular}%
}
\caption{Comparison among efficient convolutions in terms of parameter/memory compression ratio (smaller is better) and accuracy.
The cardinality $d$ in \texttt{PENNI} is 2.
\texttt{CirCNN} uses a block size $k$=4.
(\texttt{Ours}-$B_i$-$B_c$-$q_b$-$q_u$-$q_v$) is the network setup.}
\label{tab:CompareExp}
\vspace{-7pt}
\end{table*}
We perform a fine-grained investigation on the mixed-precision bitwidth $(q_b, q_u, q_v)$ to justify the trade-off between accuracy and memory efficiency.
For simplicity, we assume the same bitwidth combination for all layers.
Figure~\ref{fig:QuantCompare} plots the accuracy-memory curve with equal $q_b$, $q_u$, and $q_v$.
Above 3-bit, we can maintain over 93\% accuracy ($\sim$1\% drop).
Equal bit-precision for basis and coefficients may not be the best combination.
Thanks to our mixed-precision bit-level generation mechanism, we allow larger freedom to further explore different $q_b$, $q_u$, and $q_v$ settings around a region of interest where the accuracy starts to drop.
One key observation is that mixed-precision settings indeed can lead to higher accuracy with lower memory cost than equal settings.
We also observe that relatively-balanced settings, e.g., (2,5,3), (4,5,6), generally outperform extremely-imbalanced ones, e.g., (5,1,8), (2,4,8).
Hence, we claim that relatively-balanced mixed-precision bases are preferred to achieve better memory efficiency and less accuracy loss.

\subsection{Comparison with prior work}
\label{sec:ComparisonExp}
Our method can serve as a memory-efficient drop-in substitution for normal convolutions.
To show the superiority of our method over prior arts, we compare the memory compression ratio and inference accuracy with the baseline convolution (\texttt{Conv}) and
four representative prior works, depthwise separable Conv (\texttt{DSConv})~\cite{NN_CVPR2016_He}, single-level low-rank decomposition (\texttt{PENNI})~\cite{NN_ICML2020_Li}, blueprint Conv (\texttt{BSConv})~\cite{NN_CVPR2020_Haase}, and block-circulant Conv (\texttt{CirCNN})~\cite{NN_MICRO2017_Ding} on ResNet-18 and DenseNet-121 in Table.~\ref{tab:CompareExp}.
For fair comparisons, all methods only apply to convolutional layers and use the same training settings as mentioned.
To clarify, the selection of ($B_i$,$B_c$,$q_b$,$q_u$,$q_v$) is not from exhaustive enumeration but simply based on the target compression ratio and the heuristic design guidance we concluded.
We only evaluate the unpruned \texttt{PENNI} version since pruning is an orthogonal technique to our method.
We use a low-rank factor $d$=2 for \texttt{PENNI}~\cite{NN_ICML2020_Li} and a circulant block size $k$=4 for \texttt{CirCNN}~\cite{NN_MICRO2017_Ding} for a comparable memory cost and accuracy.

Compared with the baseline convolution, our 32-bit version achieves 5$\times$-20$\times$ memory reduction.
Compared with our special cases \texttt{DSConv} and \texttt{BSConv}, our method with a small $B_i$ and a medium $B_c$ shows 2$\times$-4$\times$ memory reduction and comparable accuracy.
Our multi-level generation outperforms the single-level low-rank decomposition method \texttt{PENNI} with 3.8$\times$-4.7$\times$ lower memory cost and better accuracy.
We outperform \texttt{CirCNN} in both metrics.
With mixed-precision generation, we boost the memory efficiency by 25$\times$-125$\times$ and 16$\times$-19$\times$ over the baseline \texttt{Conv} and the best prior work \texttt{BSConv} respectively, with competitive accuracy.
Though on DensetNet-121 CIFAR-100, we have $\sim$0.7\% accuracy drop, we have much lower memory cost.
A larger $B_c$ and higher bitwidths can be selected to recover the accuracy as a trade-off.

\subsection{Boost compact models on harder tasks}
\label{sec:DiscussMobileNet}
To fully justify our superiority, we need to answer another three important questions: 1) how does it perform on architectures that are already compact; 2) is it compatible with activation quantization that is more memory bottlenecked; and 3) does the compressed low-rank kernel still have enough representability to capture critical features in high-resolution images.
Similar to Figure~\ref{fig:KernelCorr}, we also observe strong intra-kernel correlation for depth-wise Conv (\texttt{DWConv}) and cross-kernel correlation for point-wise Conv (\texttt{PWConv}).
Hence we further apply our \textit{in-situ} generation scheme to each individual \texttt{DWConv} and \texttt{PWConv} in the inverted residual block of MobileNet-V2 for further weight compression.
Besides, we perform quantization to activation for each layer to save the most critical activation memory cost.
Table~\ref{tab:CompareMobileNetV2} shows that we can further save $>$10$\times$ weight storage and reduce the largest activation memory cost by 4$\times$ even on compact architectures.
On fine-grained image recognition tasks where the input images have high resolutions and low categorical variances, the compressed models still demonstrate strong model representability that can capture subtle but critical traits with negligible accuracy drop.
Table~\ref{tab:CompareCompactModel} evaluates our methods further on searched compact networks on detection tasks, which are known to be energy/memory-demanding, our method can lead to 5-12$\times$ compression with marginal performance loss.

\begin{table}[]
\centering
\resizebox{0.48\textwidth}{!}{%
\begin{tabular}{|l|c|c|c|c|}
\hline
\multirow{2}{*}{{\ul }} & \multicolumn{2}{c|}{CIFAR-10} & \multicolumn{2}{c|}{CIFAR-100} \\ \cline{2-5}
                        & Mem Ratio    & Acc        & Mem Ratio  & Acc        \\ \hline\hline
Original~\cite{NN_CVPR2018_Sandler} &    1.0000  & 93.06\%    &  1.0000   &   73.90\%  \\\hline
\texttt{Ours}-5-40-4-4-4                      &    0.0783  & 94.03\%    &  0.0867   &    73.11\%        \\ \hline
\texttt{Ours}-5-40-4-4-4 (A8)                    &    0.0783  & 94.02\%    &  0.0867   & 72.90\%           \\ \hline\hline
\multirow{1}{*}{{\ul }} & \multicolumn{2}{c|}{SVHN} & \multicolumn{2}{c|}{TinyImageNet-200$^{\dagger}$} \\\hline
Original~\cite{NN_CVPR2018_Sandler} &    1.0000  & 96.37\%    &  1.0000   &   67.13\%  \\\hline
\texttt{Ours}-5-40-4-4-4                      &    0.0783  & 96.61\%    &  0.1251   &    65.59\%        \\ \hline
\texttt{Ours}-5-40-4-4-4 (A8)                    &    0.0783  & 96.63\%    &  0.1251   & 65.44\%           \\ \hline\hline
\multirow{1}{*}{{\ul }} & \multicolumn{2}{c|}{StanfordDogs-120$^{\dagger}$} & \multicolumn{2}{c|}{StanfordCars-196$^{\dagger}$} \\\hline
Original~\cite{NN_CVPR2018_Sandler} &    1.0000  & 72.25\%    &  1.0000   &   89.32\%  \\\hline
\texttt{Ours}-5-40-4-4-4                      &    0.0885  & 71.06\%    &  0.0948   &    89.54\%        \\ \hline
\texttt{Ours}-5-40-4-4-4 (A8)                    &    0.0885  & 71.42\%    &  0.0948   & 89.47\%           \\ \hline
\end{tabular}%
}
\caption{\textit{In-situ} generation with activation/weight quantization on MobileNetV2~\cite{NN_CVPR2018_Sandler}.
The setup follows (\texttt{Ours}-$B_i$-$B_c$-$q_b$-$q_u$-$q_v$).
\textit{A8} means 8-bit activation.
$^{\dagger}$ means teacher models are initialized with ImageNet-pretrained models.
The setup for TinyImageNet is (6-60-5-5-5).}
\label{tab:CompareMobileNetV2}
\vspace{-5pt}
\end{table}

\begin{table}[h]
\centering
\resizebox{0.48\textwidth}{!}{%
\begin{tabular}{|l|c|c|c|c|c|c|c|}
\hline
\multirow{2}{*}{{\ul }} & \multicolumn{2}{c|}{StanfordDogs-120} & \multicolumn{2}{c|}{ImageNet-50}& \multicolumn{2}{c|}{PASCAL VOC} \\ \cline{2-7}
                        & Mem Ratio    & Acc        & Mem Ratio  & Acc   &Mem Ratio& mAP     \\ \hline\hline
MobilenetV2 (SSD-lite) &    1.0000  & 72.25\%    &  1.0000   & 87.56\%   & 1.0000 & 0.683\\\hline
\texttt{Ours} (SSD-lite)   &    0.0885  &  71.06\%  & 0.0821    & 87.52\%    & 0.1392 & 0.655       \\ \hline\hline
MobilenetV3-S (SSD-lite) &    1.0000  & 65.41\%    &  1.0000   & 85.04\%    & 1.0000 & 0.544\\\hline
\texttt{Ours} (SSD-lite)                      &    0.2082  &  66.64\%   &  0.2060   &  85.44\%   & 0.2238 & 0.513      \\ \hline\hline
EfficientNet-B0 &    1.0000  & 75.43\%    &  1.0000   & 89.56\%    & - & - \\\hline
\texttt{Ours}                    &    0.1257  & 75.00\%    &  0.1132  & 88.52\% & - & - \\\hline
\end{tabular}%
}
\caption{Evaluate compact models beyond simple tasks and  classification.}
\label{tab:CompareCompactModel}
\vspace{-10pt}
\end{table}

\vspace{-.05in}
\section{Conclusion}
\label{sec:Conclusion}
In this work, we propose a general and unified framework for memory-efficient DNN designs via multi-level \textit{in situ} generation.
We jointly leverage the intrinsic correlation and bit-level redundancy within convolutional kernels and allow the ultra-fast accelerator to generate the weights \textit{in situ} by itself to boost the performance.
A photonic case study is given to show our latency/power advantages.
Experiments show that our method achieves 10$\times$-20$\times$ memory efficiency boost compared with prior methods.
Our method provides a unified view to prior single-level low-rank methods and enables a new design paradigm to break through the ultimate memory bottleneck for emerging DNN accelerators by their tremendous computing power.

\section*{Acknowledgment}
The authors acknowledge the Multidisciplinary University Research Initiative (MURI) program through the Air Force Office of Scientific Research (AFOSR), contract No. FA 9550-17-1-0071, monitored by Dr. Gernot S. Pomrenke.

\end{document}